\documentclass{IEEEtran}
\usepackage{times}
\usepackage{epsfig}
\usepackage{graphicx}
\usepackage{amsmath}
\usepackage{amssymb}
\usepackage{bm}
\usepackage{multirow}
\usepackage{makecell}
\usepackage[pagebackref=true,breaklinks=true,colorlinks,bookmarks=false]{hyperref}

\begin{document}

\title{Memory Matters: Convolutional Recurrent Neural Network for Scene Text Recognition}

\author{%
    \IEEEauthorblockN{%
        Qiang Guo\IEEEauthorrefmark{1}, Dan Tu, Guohui Li and Jun Lei\\
    }
    \IEEEauthorblockA{Department of Information System and Management\\
    National University of Defense Technology\\
    Email: \{guoqiang05\IEEEauthorrefmark{1}, tudan, guohuili, junlei\}@nudt.edu.cn
    }
}

\IEEEpubid{}

\maketitle

\begin{abstract}
Text recognition in natural scene is a challenging problem due to the many factors affecting text appearance.
In this paper, we presents a method that directly transcribes scene text images to text without needing of sophisticated character segmentation. 
We leverage recent advances of deep neural networks to model the appearance of scene text images with temporal dynamics.
Specifically, we integrates convolutional neural network (CNN) and recurrent neural network (RNN) which is motivated by observing the complementary modeling capabilities of the two models. 
The main contribution of this work is investigating how temporal memory helps in an segmentation free fashion for this specific problem.
By using long short-term memory (LSTM) blocks as hidden units, our model can retain long-term memory compared with HMMs which only maintain short-term state dependences. We conduct experiments on Street View House Number dataset containing highly variable number images. The results demonstrate the superiority of the proposed method over traditional HMM based methods.
\end{abstract}

\section{Introduction}\label{sec:intro}

Text recognition in natural scene is an important problem in computer vision. However, due to the enormous appearance variations in natural images, e.g. different fonts, scales, rotations, illumination conditions, it is still quite challenging.

Identifying the position of a character and recognizing it are two interdependent problems. Straight-forward methods treat the task as separate character segmentation and recognition\cite{Bissacco2013,netzer2011reading}. This paradigm is fragile in unconstrained natural images for it's  difficult to deal with low resolution, low contrast, blurring, large diversity of text fonts and highly complicated background clutters. 


Due to the shortcoming of these methods, algorithms combining segmentation and recognition were proposed. GMM-HMMs are the mostly used models, especially in speech and handwriting communities\cite{bourlard2012connectionist,dahl2012context,marti2001using,espana2011improving}. 

In this paradigm, scene text images are transformed to frame sequences by sliding window. GMMs are used for modeling frame appearance and HMMs are used to infer the target labels of the whole sequence.\cite{Guo2015}

The merit of this method is avoiding the need of fragile character segmentation. However HMMs have several obvious shortcomings, e.g. lacking of long context consideration, improper independent hypothesis etc.

\begin{figure}[!t]
  \begin{center}
    \includegraphics[width=0.8\columnwidth]{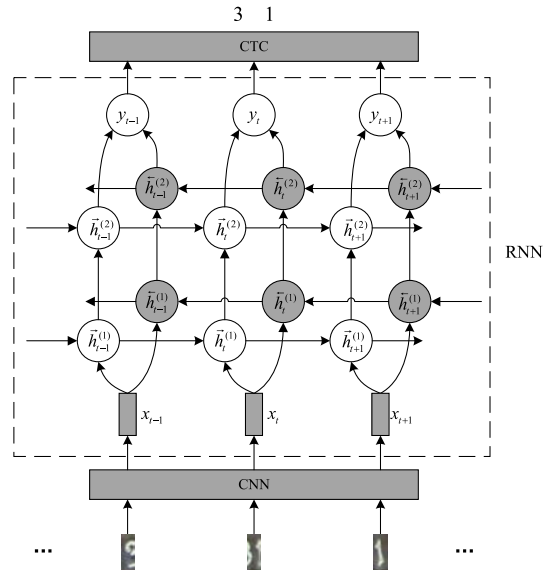}
  \caption{The whole architecture of CRNN.}\label{fig:crnn}
  \end{center}
\end{figure}

As the developments of deep neural networks (DNNs) flourishing, convolutional neural networks (CNNs) have been used to form the hybrid CNN-HMM model\cite{Guo2015}, replacing GMMs as the observation model. The model generally performs better than the GMM-HMM model thanks to the strong representation capacity of CNN, however still doesn't eliminate the issues with HMM.

\IEEEpubidadjcol

In this work, we address the issues of HMMs while keeping the algorithm free of segmentation. The novalty of our method is using Recurrent Neural Network (RNN), which has the ability of adaptively retaining long-term dynamic memory, as the sequence model. We combine CNN with RNN to utilize their representation abilities on different aspects.

RNN is a powerful connectionist model for sequences. Comparing with static feed-forward networks, it introduces recurrent connections enabling the network to maintain an internal state. It doesn't make any hypothesis on the independence of inputs, so each hidden unit can take into account more input information. Specifically, LSTM memory blocks are used enabling RNN to retain longer range of inter-dependences of input frames. Another virtue of using RNN as the sequence model is the ease of build an end-to-end trainable system directly trained on the whole image without needing explicit segmentation. The main weakness of RNN is its feature extraction capability.

To alleviate the shortcomings of both HMM and RNN, we propose a novel end-to-end sequence recognition model named Convolutional Recurrent Neural Network (CRNN). The model is composed with hierarchical convolutional feature extraction layers and recurrent sequence modeling layers.

CNN is good at appearance modeling and RNNs have strong capacity for modeling sequences. The model is trained with Connectionist Temporal Classification (CTC)\cite{Graves2006} object which enables the model directly learned from images without segmentation information. 

Our idea is motivated by observing the complementary modeling capacities of CNN and RNN, and inspired by recent success applications of LSTM architectures to various sequential problems, such as handwriting\cite{Graves2008}, and speech recognition\cite{Graves2014}, image description\cite{Donahue2014,Karpathy2014}. The whole architecture of our model is illustrated in \autoref{fig:crnn}. 

\section{Related Work}
In this section, we briefly survey methods that focus on sequence modeling without segmentation. 

The paradigms of scene text recognition algorithms are similar with handwriting recognition. Straightforward methods\cite{Bissacco2013,netzer2011reading} are composed of two separated parts. A segmentation algorithm followed by a classifier to determine the category or each segment. Often, the classification results are post-processed to form the final results.

To eliminate the need of explicit character segmentation, many researchers use GMM-HMM for text recognition\cite{Vinciarelli2004,Kozielski2013}. GMM-HMM is a classical model widely used by the speech community. HMMs make it possible to model sequences without the necessity of segmentation. However, there are many shortcomings of GMM-HMM, which make it not widely used in scene text recognition. Firstly, GMM is a weak model for modeling characters in natural scene. Secondly, HMM has many limitations that are addressed in \autoref{sec:intro}.

To strengthen the representation capability of HMM based model, CNN is then used to replace GMM as the observation model which forms the hybrid CNN-HMM model\cite{Guo2015,Bluche2013a}. While improves the performance in comparison with GMM-HMM, it still doesn't eliminate the shortcomings of HMM.

Our idea is motivated by recent success of the RNN models applied to handwriting recognition\cite{Graves2008}, speech recognition\cite{Graves2014} and image description\cite{Donahue2014,Karpathy2014}. The main inspiration of our idea is observing the complementary modeling capacity of CNN and RNN. CNN can automatically learn hierarchical image features but only as a static model. RNN is good at sequence modeling while lacking the ability of feature extraction. We integrate the two models to form an end-to-end scene text recognition system. 

Different with recent works\cite{Elagouni2012} which use similar ideas, we investigate to use deep RNNs by stacking multiple recurrent hidden states on top of each other. Our experiment shows the improvements of the endeavor.

\section{Problem Formulation}
We formulate scene text recognition as a sequence labeling problem by treating scene text as frame sequences. The label sequence is drawn from a fixed alphabet $L$. The length of the label sequence is not restricted to be equal to that of the frame sequence. 

Each scene text image is treated as a sequence of frames denoted by ${\bm x}=(x_1, x_2, \cdots, x_T)$. The target sequence is ${\bm z}=(z_1, z_2, \cdots, z_U)$. We use bold typeface to denote sequences. 

We constrain that $|{\bm z}|=U\le |{\bm x}|=T$. The input space $\mathcal{X}=(\mathbb{R}^M)^*$ is the set of all sequences of $M$ real valued vectors. The target space $\mathcal{Z}=L^*$ is the set of all sequences over the alphabet $L$ of labels. We refer ${\bm z}\in L^*$ as a {\it labeling}.

Let $S$ be a set of training samples drawn independently from a fixed distribution $\mathcal{D_{X\times Z}}$ composed of sequence pairs $({\bm x}, {\bm z})$. The task is to use $S$ to train a sequence labeling algorithm $f:\mathcal{X}\mapsto\mathcal{Z}$ to label the sequences in a test set $S'\in \mathcal{D_{X\times Z}}$ as accurately as possible given the error criterion {\it label error rate} $E^{lab}$:
\begin{equation}
  E^{lab}(h, S')=\frac 1Z \sum_{({\bm x, z})\in S'} ED(h({\bm x}), {\bm z})
\end{equation}
where $ED({\bm p, \bm q})$ is the {\it edit distance} between two sequences ${\bm p}$ and ${\bm q}$.

\section{Method}

\subsection{The proposed model}
The network architecture of our CRNN model is shown in \autoref{fig:crnn}. The model is mainly composed with two parts, a deep convolutional network for feature extraction and a bidirectional recurrent network for sequence modeling. 

An scene text image is transformed into a sequence of frames which are fed into the CNN model to extract feature vectors. 

The CNN model only map one frame feature to its corresponding output vector. The sequence of feature vectors are then used as the input of RNN which takes the whole sequence history into consideration. 

The recurrent connections allow the network to retain previous inputs as memory in the internal states and discovery temporal correlations among time-steps even far from each other. 

Given an input sequence ${\bm x}$, a standard RNN computes the hidden vector sequence ${\bm h}=(h_1, h_2, \cdots, h_T)$ and output vector sequence ${\bm y}=(y_1, y_2, \cdots, y_T)$ as following:
\begin{align}
  h_t &= \mathcal{H}(W_{ih}x_t+W_{hh}h_{t-1}+b_h) \\
  y_t &= W_{ho}h_t+b_o
\end{align}
where the $W$ terms denote weight matrices (e.g. $W_{ih}$ is the input-hidden weight matrix), the $b$ terms denote bias vectors (e.g $b_h$ is hidden bias vector), $\mathcal{H}$ is the hidden layer activation function.

We stack a CTC layer on top of RNN. With the CTC layer, we can train the RNN model directly on the sequences' labellings with knowing frame-wise labels. 

\subsection{Feature extraction}

CNNs\cite{Lecun1998,Krizhevsky2012} have shown exceptionally powerful representation capability for images and have achieved state-of-the-art results in various vision problems. In this work, we build an CNN for feature extraction. 

We use CNN as a transforming function $f({\bm o})$ that takes an input image ${\bm o}$ and outputs an fixed dimensional vector $x$ as the feature. The convolution and pooling operations in deep CNNs are specially designed to extract visual features hierarchically, from local low-level features to robust high-level ones. The hierarchically extracted features are robust to variable factors that characters facing in natural scene.

\subsection{Dynamic modeling with Bidirectional RNN and LSTM}
One shortcoming of conventional RNNs is that they only able to make use of previous context. However, it's reasonable to exploit both previous and future contexts. For scene text recognition, the left and right context are both useful for determining the category of a specific frame image.

In our model, we use Bidirectional RNN (BRNN)\cite{Schuster1997} to process sequential data from both directions with two separate hidden layers.

BRNN computes the {\it forward} hidden sequence $\overrightarrow{\bm h}$, the {\it backward} hidden sequence $ \overleftarrow{\bm h} $. Each time-step $y_t$ of the output sequence is computed by integrating both directional hidden states:
\begin{equation}
  y_t = W_{\overrightarrow{h}y}\overrightarrow{h}_t+W_{\overleftarrow{h}y}\overleftarrow{h}_t+b_o
\end{equation}

BRNN provides the output layer with complete past and future context for every time-step in the input sequence. 
During the forward pass, input sequence is fed to both directional hidden layers, and the output layer is not updated until both the two hidden layers have processed the entire input sequence.
The backward pass of BPTT for BRNN is similar with unidirectional RNN, except that the output layer error $\delta$ is fed back to the both two directional hidden layers.

While in principle RNN is a simple and powerful model, in practice, it's unfortunately hard to train properly. RNN can be seen as a deep neural network unfolding in time. A critical problem when training deep networks is the {\it vanishing and exploding gradient} problem\cite{Pascanu2013}. When error signal transmitting along the network's recurrent connections, it decays or blows up exponentially. Due to this, the range of context being accessed can be quite limited.

Long Short-term Memory (LSTM) block\cite{Hochreiter1997} is designed to control the input, output and transition of signal so as to retain useful information and discard useless one. LSTMs are used as memory blocks of RNN and can alleviate the vanishing and exploding gradient issues. They use a special structure to form memory cells. The LSTM updates for time-step $t$ given inputs $x_t$, $h_{t-1}$ and $c_{t-1}$ are:
\begin{align}
  i_t &= \sigma(W_{xi}x_t+W_{hi}h_{t-1}+W_{ci}c_{t-1}+b_i)\\
  f_t &= \sigma(W_{xf}x_t+W_{hf}h_{t-1}+W_{cf}c_{t-1}+b_f)\\
  c_t &= f_{t}c_{t-1}+i_{t}tanh(W_{xc}x_t+W_{hc}h_{t-1}+b_c)\\
  o_t &= \sigma(W_{xo}x_t+W_{ho}h_{t-1}+W_{co}c_{t}+b_o)
\end{align}
where $\sigma$ is the logistic sigmoid function, $i,f,o,c$ are respectively the {\it input gate, forget gate, output gate} and {\it cell} activation vectors, all of which are the same size as the hidden vector $h$.

The core part of LSTM block is the memory cell $c$ that encodes the information of the inputs that have been observed up to that step. The gates determine whether the LSTM keeps the value from the gate or discards it. The input gate controls whether the LSTM considers its current input, the forget gate allows to forget its previous memory, and the output gate decides how much of the memory to transfer to the hidden state. Those features enable the LSTM architecture to learn complex long-term dependences.

\subsection{Training}

Both the CNN and RNN models are deep models. When stacking them together, it is difficult to train them together. Starting with an random initialization, the supervision information propagated from RNN to CNN are quite ambiguous. So we separately train the two parts.

The CNN model is trained with stochastic gradient descent. The samples for training CNN is got by performing forced-alignment on the scene text images with a GMM-HMM model\cite{Guo2015}. 


For training the RNN model, we need a loss function that can directly compute the probability of the target labelling from the frame-wise outputs of RNN given the observations.

Connectionist Temporal Classification (CTC)\cite{Graves2006} is an objective function designed for sequence labeling problem when the segmentation of data is unknown. It does not require pre-segmented training data, or post-processing to transform the network outputs into labelings. It trains the network to map directly from input sequences to the conditional probabilities of the possible labelings.

A CTC output layer contains one more unit than there are elements in the alphabet $L$, denoted as $L'=L\cup\{\mathtt{nul}\}$. The elements in $L'^*$ are refered as {\it paths}. For an input sequence $\bm x$, the conditional probability of a path $\pi\in L'^T$ is given by
\begin{equation}
  p(\pi|{\bm x})=\prod_{t=1}^{T}y_{\pi_t}^t
\end{equation}
where $y_k^t$ is the activation of output unit $k$ at time $t$. An operator $\mathcal{B}$ is defined to merge the repeated labels and remove blanks. For example, $\mathcal{B}(-,3,3,-,-,3,2,2)$ yields the labeling $(3,3,2)$. The conditional probability of a given labeling ${\bm l}$ is the sum of the probabilities of all paths corresponding to it:
\begin{equation}
  p({\bm l}|{\bm x})=\sum_{\pi\in\mathcal{B}^{-1}(\bm l)}p(\pi|{\bm x}).
\end{equation}

We use CTC\cite{Graves2006} as the objective function. A {\it forward-backward algorithm}\cite{Graves2006} for CTC, which is similar to the forward-backward algorithm of HMM, is designed to effectively evaluate the probability.

The objective function for CTC is the negative log probability of the correct labelings for the entire training set:

\begin{equation}
  O^{CTC}=-\sum_{({\bm x,z})\in S}\ln(p({\bm z}|{\bm x})).
\end{equation}

Given the partial derivatives of some differential loss function $\mathcal{L}$ with respect to the network outputs, we use {\it back propagation through time algorithm} (BPTT) to determine the derivatives with respect to the weights. 

Like standard {\it back propagation}, BPTT follows the chain rule to calculate derivatives. The subtle difference is that the loss function depends on the activation of the hidden layer not only through its influence on the output layer, but also through the hidden layer of next time-step. So the error back propagation formula is
\begin{equation}
  \delta_h^t=\theta'(a_h^t)\left(\sum_{k=1}^K \sigma_k^t w_{hk}+\sum_{h'=1}^H \sigma_{h'}^{t+1} w_{hh'}\right)
\end{equation}
where
\begin{equation}
  \delta_j^t\equiv\frac{\partial \mathcal{L}}{\partial a_j^t}
\end{equation}

The same weights are reused at every timestep, so we sum the whole sequence to get the derivatives with respect to the network weights:
\begin{equation}
  \frac{\partial \mathcal{L}}{\partial w_{ij}}=\sum_{t=1}^{T}\frac{\partial\mathcal{L}}{\partial a_j^t}\frac{\partial a_j^t}{\partial w_{ij}}=\sum_{t=1}^T \delta_j^t b_j^t
\end{equation}

\subsection{Decoding}

The {\it decoding} task is to find the most probable labeling ${\bm l}^*$ given an input sequence ${\bm x}$:
\begin{equation}
  \label{equ:decoding}
  {\bm l}^*=\arg\max_{\bm l}p({\bm l}|{\bm x}).
\end{equation}

We use a simple and effective approximation by choosing the most probable path, then get the labeling ${\tilde{\bm l}}^*$ corresponding to the path:
\begin{equation}
  {\tilde{\bm l}}^*=\mathcal{B}(\arg\max_{\pi}p(\pi|{\bm x})).
\end{equation}

\section{Experiments}
This section presents the details of our experiments. We compare the CRNN model with $3$ methods: (1) A baseline method by directly using CNN to predict the character at each time-step, with a post-processing procedure to merge repeated outputs; (2) The GMM-HMM model; (3) Hybrid CNN-HMM model.

\subsection{Dataset}
We explore the use of CRNN model on a challenging scene text dataset Street View House Numbers(SVHN)\cite{netzer2011reading}. The dataset contains two versions of sample images. One contains only isolated $32\times32$ digits with $73257$ for training and $26032$ for testing. The other contains unsegmented full house number images containing variable number of unsegmented digits. The full number version is composed of $33402$ training images and $13068$ testing.
House numbers in the dataset show quite large appearance variability, blurring and unnormalized layouts. 

We use the isolated version of the dataset for training the CNN, which is then used to extract features for each sequence frame. The full house number version is used for HMM based methods and CRNN. 

The training samples are randomly splited out $10\%$ as the validation set. The validation set is used only for tuning hyper parameters of different models. All the models use the same training, validation and testing set, which makes it fair to compare different models. 

\subsection{Implementation details}
The full number images are normalized to the same height $32$ while keeping the scale ratio, then transformed to frame sequences by $32\times20$ sliding window. Each frame $o_t$ is fed into CNN to producing feature $x_t=f(o_t)$. 
We standardize the features by subtracting the global mean and dividing the standard deviation.

Our CNN model contains $3$ convolutioin and pooling layer pairs and $2$ fully connected layers. The activation neurons are all rectified linear units (ReLU)\cite{Nair2010}. The output number sequence of the layers are $32,32,64,128,10$. CNN is trained with stochastic gradient descent (SGD) under cross entropy loss with learning rate $10^{-3}$ and momentum $0.9$. 

We choose the output of CNN's first fully connected layer as frame features. The features are $128$D and used for both HMM based models and CRNN.

We use $11$ HMMs coinciding with the extended alphabet $L'$. All the HMMs are of $3$-state left-to-right topology, except for the $\mathtt{nul}$ category which has $1$ self-looped state. The GMM-HMM model is trained with Baum-Welch algorithm.

For the proposed CRNN model, we use a deep bidirectional RNN. We stack $2$ RNN hidden layers, both of which are bidirectional. All hidden units are LSTM blocks. 
The CRNN model is trained with BPTT algorithm using SGD. We use a learning rate of $10^{-4}$ and a momentum of $0.9$.

\subsection{Results}
\begin{figure}[!ht]
  \begin{center}
    \includegraphics[width=0.7\columnwidth]{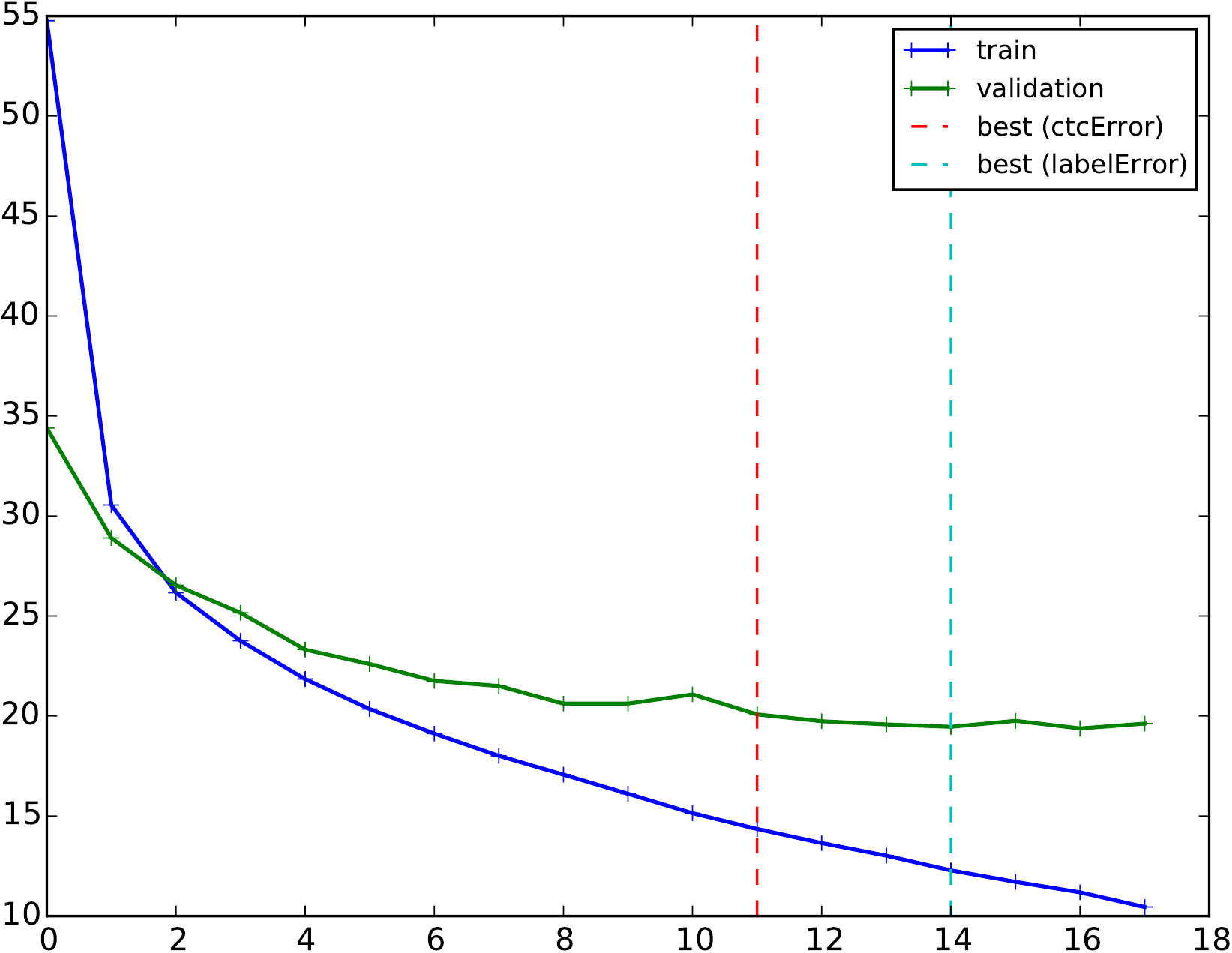}    
  \end{center}
  \caption{Training curve during training of CRNN-2-layers.}\label{fig:curve}
\end{figure}

\paragraph{CNN based sequence labelling} 
We train the CNN on the isolated version of SVHN, then use it to get the character predictions for each frame, choosing the most probable one as the result. After that, we merge consecutively repeated characters to get the final sequence labels. 
The recognition accuracy on the full number test set is only $0.23$. Note that, the CNN model achieve an accuracy of $0.92$ on the isolated test set. The correctly recognized house numbers by CNN are mostly contain only $1$ or $2$ digits. The simple experiment can gain us an intuition of how hard the problem is and how important the sequential information is.

\paragraph{HMM based models}
We compare our method with $2$ kinds of HMM based models. One is GMM-HMM model, the other is hybrid CNN-HMM model\cite{Guo2015}. 
The number of mixture components is an important factor for GMM-HMM. We evaluate different number of Gaussian mixture components. The sequence accuracy stops improving at $800$. The model tends to overfit when we continually increase the number of mixture components.

Hybrid CNN-HMM improves GMM-HMM by using CNN as the observation model. The training process is an iterative procedure, where network retraining is alternated with HMM re-alignment to generate more accurate state assignments. 

\begin{table}[t]
  \begin{center}
    \begin{tabular}{c|c|c|c|c}
      \Xhline{1pt}
      \multirow{2}*{} & \multicolumn{2}{c|}{CRNN-1-layer} & \multicolumn{2}{c}{CRNN-2-layers} \\
      \Xcline{2-5}{0.5pt}
      & epoch & accuracy & epoch & accuracy \\
      \Xhline{0.5pt}
      CTC error & 12 & 0.84 & 9 & 0.90 \\
      label error & 15 & 0.86 & 10 & 0.91 \\
      \Xhline{1pt}
    \end{tabular}  
  \end{center}
  \caption{Comparison of different CRNN architectures.}\label{tbl:archs}
\end{table}

\paragraph{CRNN}
Recent developments of deep learning shows that \textit{Deep} is an important factor for feed-forward models to gain stronger representation capability\cite{pascanu2013number}. We evaluated two architectures of CRNN. One uses $1$ hidden layer denoted as CRNN-1-layer, the other $2$ hidden layers denoted as CRNN-2-layers. CRNN-1-layer has $128$ LSTM memory cells, CRNN-2-layers has $128$ for the first hidden layer and $32$ for the second. \autoref{fig:crnn} shows the architecture of CRNN-2-layers.

Experiment results are presented in \autoref{tbl:archs}. \textit{Epoch} column lists the epoch at which the best model reaches with respect to different error criterion. \textit{Accuracy} column shows the sequence accuracy of the best model on test set.

As can be seen, the deeper architecture performs not only better but also with less training epochs. This is a surprising finding, as intuitively the deeper model has more parameters which makes it more difficult to train.

\begin{table}[!th]
  \begin{center}
  \begin{tabular}{c|c}
    \Xhline{1pt}
    Model & Accuracy \\
    \Xhline{0.5pt}
    CNN & 0.23\\
    GMM-HMM & 0.56\\
    Hybrid CNN-HMM & 0.81 \\ 
    CRNN & \textbf{0.91}\\
    \Xhline{1pt}
  \end{tabular}
  \end{center}
  \caption{Performance comparison of different models.}\label{tbl:models}
\end{table}

Performance comparison of CRNN with other models is represented in \autoref{tbl:models}. As shown by the experiments, CRNN outperforms CNN and both HMM based models.

\section{Conclusion}
We have presented the Convolutional Recurrent Neural Network (CRNN) model for scene text recognition. It uses CNN to extract robust high-level features and RNN to learn sequence dependences. The model eliminates the need of character segmentation when doing scene text recognition. We apply our method on street view images and achieve promising results. CRNN performs much better than HMM based methods. However, CNN is still trained separately. While the recognition process is segmentation-free, we still need cropped character samples for training the CNN. To eliminate the needing of cropped samples in training, we plan to investigate using forced alignment of GMM-HMM for bootstrapping of CRNN. Better method would be directly perform joint training of CNN and RNN from scratch. Another promising direction would be to investigate the potential of stacking more hidden LSTM layers of RNN.

\section*{Acknowledgment}
This work was supported by Open Project Program of the State Key Laboratory of Mathematical Engineering and Advanced Computing(No.2015A04).

\bibliographystyle{IEEEtran}
\bibliography{ref}

\end{document}